\documentclass[11pt]{article}
\ifdefined\REVIEW
  \usepackage[review]{acl}
\else
  \usepackage[]{acl}
\fi
\newcommand{\codeavail}{\ifdefined\REVIEW the supplementary material\else \url{https://github.com/vedant-jumle/Seahorse}\fi}
\usepackage{times}
\usepackage{latexsym}
\usepackage[T1]{fontenc}
\usepackage[utf8]{inputenc}
\usepackage{microtype}
\usepackage{graphicx}
\usepackage{url}
\usepackage{amsmath,amssymb}
\usepackage{booktabs}
\usepackage{xcolor}
\usepackage{pifont}
\usepackage{tikz}
\usepackage{placeins}
\usetikzlibrary{arrows.meta,positioning,calc}

\title{What Survives When Activation Steering Is Used as Memory}

\author{Vedant Jumle \\
  Delft University of Technology \\
  \texttt{vvjumle@student.tudelft.nl}}

\begin{document}
\maketitle

\begin{abstract}
Activation steering has recently been used as a memory for skills: past experiences of an agent are stored as steering vectors and retrieved later. Facts and preferences about the user are assumed to need text instead. We test that assumption with a training-free activation memory beside a frozen LLM. The change an experience causes in late-layer states is stored as a shift, filed under a whitened key for the conversation's gist, and added back in a later session only when the gist matches. We argue that such a shift is a fixed average of an effect that attention, in context, recomputes for every question, so only question-independent content can survive. In a pre-registered study on Qwen3.5-2B and 9B, with judges from four model families, the memory is selective (about 98\% of unrelated replies are unchanged), but facts are stored as the word to say, in the last layers only, and are almost never used (at most 15\% clean use, against 96--98\% in context). Leanings transfer only in a narrow strength window. Most strikingly, the reference subtracted when writing a preference decides whether its concept or its direction survives: unless the opposite is subtracted, a stored dislike is recalled as a like.
\end{abstract}

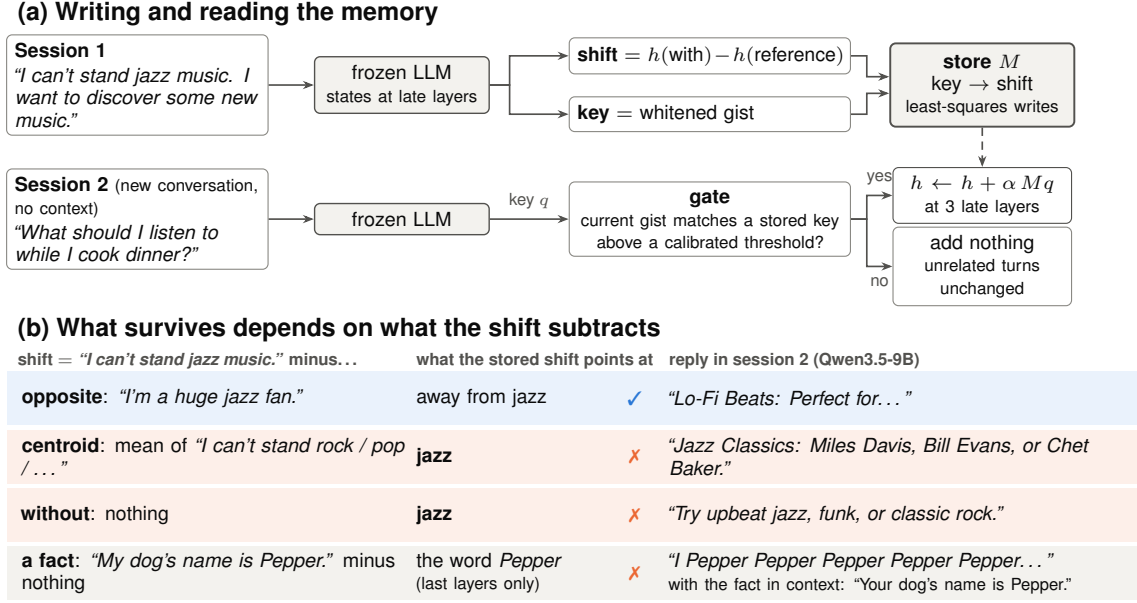
\begin{figure*}[t]
  \centering
  \resizebox{\textwidth}{!}{
\definecolor{sBlue}{HTML}{2A78D6}
\definecolor{sOrange}{HTML}{EB6834}
\definecolor{sInk}{HTML}{0B0B0B}
\definecolor{sInk2}{HTML}{52514E}
\definecolor{sMuted}{HTML}{8A8984}
\definecolor{sFill}{HTML}{F3F2EE}
\definecolor{sBlueBg}{HTML}{EAF2FC}
\definecolor{sOrangeBg}{HTML}{FDEFE9}
\begin{tikzpicture}[
  font=\sffamily\scriptsize, text=sInk,
  box/.style={draw=sMuted, rounded corners=2pt, fill=white, inner sep=3pt, align=left, line width=0.4pt},
  model/.style={draw=sInk2, rounded corners=2pt, fill=sFill, inner sep=3pt, align=center, line width=0.5pt},
  store/.style={draw=sInk2, rounded corners=2pt, fill=sFill, inner sep=4pt, align=center, line width=0.7pt},
  arr/.style={-{Stealth[length=4pt]}, line width=0.6pt, draw=sInk2},
  lbl/.style={font=\sffamily\tiny, text=sInk2, align=center},
  head/.style={font=\sffamily\bfseries\footnotesize, anchor=west},
]
\node[head] at (0, 3.15) {(a) Writing and reading the memory};

\node[box, text width=3.1cm] (exp) at (1.65, 2.25)
  {\textbf{Session 1}\\[1pt]\textit{``I can't stand jazz music. I want to discover some new music.''}};
\node[model, text width=2.0cm] (m1) at (5.0, 2.25) {frozen LLM\\{\tiny states at late layers}};
\node[box, text width=3.35cm] (shift) at (8.9, 2.62)
  {\textbf{shift} $=h(\text{with}) - h(\text{reference})$};
\node[box, text width=3.35cm] (key) at (8.9, 1.88)
  {\textbf{key} $=$ whitened gist};
\node[store, text width=2.05cm] (st) at (12.35, 2.25) {\textbf{store} $M$\\key $\to$ shift\\{\tiny least-squares writes}};
\draw[arr] (exp) -- (m1);
\draw[arr] (m1.east) -- ++(0.25,0) |- (shift.west);
\draw[arr] (m1.east) -- ++(0.25,0) |- (key.west);
\draw[arr] (shift.east) -- ++(0.2,0) |- ([yshift=3pt]st.west);
\draw[arr] (key.east) -- ++(0.2,0) |- ([yshift=-3pt]st.west);

\node[box, text width=3.1cm] (q) at (1.65, 0.55)
  {\textbf{Session 2} {\tiny (new conversation, no context)}\\[1pt]\textit{``What should I listen to while I cook dinner?''}};
\node[model, text width=2.0cm] (m2) at (5.0, 0.55) {frozen LLM};
\node[box, text width=3.35cm, align=center] (gate) at (8.9, 0.55)
  {\textbf{gate}\\{\tiny current gist matches a stored key}\\{\tiny above a calibrated threshold?}};
\node[box, text width=2.05cm, align=center, draw=sInk2] (add) at (12.35, 0.85) {$h \leftarrow h + \alpha\, M q$\\{\tiny at 3 late layers}};
\node[box, text width=2.05cm, align=center] (none) at (12.35, -0.05) {add nothing\\{\tiny unrelated turns unchanged}};
\draw[arr] (q) -- (m2);
\draw[arr] (m2) -- node[lbl, above] {key $q$} (gate);
\draw[arr] (gate.east) -- ++(0.2,0) |- node[lbl, pos=0.75, above] {yes} (add.west);
\draw[arr] (gate.east) -- ++(0.2,0) |- node[lbl, pos=0.75, below] {no} (none.west);
\draw[arr, densely dashed] (st.south) -- (add.north);

\node[head] at (0, -0.85) {(b) What survives depends on what the shift subtracts};
\node[lbl, anchor=west, font=\sffamily\tiny\bfseries] at (0.0, -1.25) {shift $=$ \textit{``I can't stand jazz music.''} minus\ldots};
\node[lbl, anchor=west, font=\sffamily\tiny\bfseries] at (5.05, -1.25) {what the stored shift points at};
\node[lbl, anchor=west, font=\sffamily\tiny\bfseries] at (8.25, -1.25) {reply in session 2 (Qwen3.5-9B)};

\newcommand{\brow}[6]{
  \fill[#2] (0,#1-0.34) rectangle (14.3,#1+0.34);
  \node[anchor=west, text width=4.85cm] at (0.05,#1) {#3};
  \node[anchor=west, text width=2.65cm] at (5.05,#1) {#4};
  \node[anchor=center] at (7.95,#1) {#5};
  \node[anchor=west, text width=5.95cm] at (8.25,#1) {\textit{#6}};
}
\brow{-1.72}{sBlueBg}{\textbf{opposite}: \textit{``I'm a huge jazz fan.''}}
     {away from jazz}{\textcolor{sBlue}{\ding{51}}}{``Lo-Fi Beats: Perfect for\ldots''}
\brow{-2.46}{sOrangeBg}{\textbf{centroid}: mean of \textit{``I can't stand rock / pop / \ldots''}}
     {\textbf{jazz}}{\textcolor{sOrange}{\ding{55}}}{``Jazz Classics: Miles Davis, Bill Evans, or Chet Baker.''}
\brow{-3.20}{sOrangeBg}{\textbf{without}: nothing}
     {\textbf{jazz}}{\textcolor{sOrange}{\ding{55}}}{``Try upbeat jazz, funk, or classic rock.''}
\brow{-3.94}{sFill}{\textbf{a fact}: \textit{``My dog's name is Pepper.''} minus nothing}
     {the word \textit{Pepper}\\{\tiny (last layers only)}}{\textcolor{sOrange}{\ding{55}}}{``I Pepper Pepper Pepper Pepper Pepper\ldots''\\{\upshape\tiny with the fact in context: ``Your dog's name is Pepper.''}}
\end{tikzpicture}}
  \caption{\textbf{(a)} A training-free activation memory beside a frozen LLM. In session~1 the change an experience
  causes in late-layer states is stored as a \emph{shift}, filed under a whitened \emph{key} for the conversation's gist.
  In a later session the shift is added back only where the current gist matches a stored key.
  \textbf{(b)} What the stored shift carries depends on the reference subtracted when writing it. Subtracting the
  opposite keeps the \emph{direction}; subtracting alternatives or nothing keeps the \emph{concept}, so a dislike is
  recalled as a like. A fact is stored as the word to say. Replies are single real samples from Qwen3.5-9B (preferences
  $\alpha{=}1$, fact $\alpha{=}2$); quantitative results are in \S\ref{sec:results}.}
  \label{fig:main}
\end{figure*}

\section{Introduction}
\label{sec:intro}

An assistant that is told \textit{``I can't stand jazz''} on Monday should not suggest a jazz playlist on Friday. Language models keep such information in one of two places. They keep it in their weights, which change only through training or model editing \citep{meng2022rome,meng2023memit}, or in their context, which is exact but is lost when the session ends unless it is retrieved and re-inserted \citep{lewis2020rag,packer2023memgpt}. A third place has recently been proposed: the model's \emph{activations}. Activation steering adds a vector to the residual stream at inference time and reliably shifts dispositions such as honesty, refusal or persona \citep{turner2023actadd,zou2023repe,rimsky2024caa,lee2025cast,cao2024bipo}. \citet{zhao2026npm} go one step further and store an agent's past experience as retrieved steering vectors, which works well as \emph{procedural} memory, i.e.\ for skills. They explicitly leave \emph{declarative} content (facts and preferences about the user) to text retrieval, on the assumption that it does not belong in activations.

That assumption has not been tested, and it matters for personalisation. Steering-based personalisation either keeps the user's context in the prompt \citep{he2025cos,liu2024icv} or trains a vector for a behaviour chosen in advance \citep{cao2024bipo}. Neither shows what happens when what a user said in one conversation is stored as an activation change and replayed in a later conversation that no longer contains it.

We build such a memory and measure what survives. It is training-free and sits beside a frozen model (Figure~\ref{fig:main}a). When an experience occurs, the change it causes in late-layer states is stored as a \emph{shift}, filed under a whitened \emph{key} for the gist of the conversation. In a later session the shift is added back only where the current gist matches a stored key. Mathematically, the store is the linear associative memory that ROME and MEMIT write into MLP weights \citep{meng2022rome,meng2023memit}, kept in the activations instead, and the gate is a training-free variant of conditional steering \citep{lee2025cast}.

Our account of what this memory can hold is simple. In context, attention recomputes an experience's effect for each new question \citep{liu2024icv,geva2023dissecting}. A stored shift is one \emph{fixed average} of that effect, so only the part of the experience that is the same for every later question can survive. This yields three predictions, which we test in a pre-registered study on Qwen3.5-2B and Qwen3.5-9B, with labels from a panel of LLM judges from different model families:
\begin{itemize}
  \item \textbf{Facts fail.} A fact matters for one question, so its averaged shift reduces to ``say the word''. The stored shift ranks the fact's word in the top~10 of the vocabulary only in the last sixth of the network. Replies that use the fact cleanly stay at or below 15\% at every strength, against 96--98\% with the fact in context. Recall appears instead as loops and garbled text, and the memory never serves as a premise for yes/no questions.
  \item \textbf{Leanings survive only in a narrow window.} Broad preferences shift open-ended replies at moderate strength, but coherence collapses as strength grows. This claim failed its pre-registered test and is sensitive to the choice of judge.
  \item \textbf{What the shift subtracts decides what is stored.} A shift is the state with the experience minus a reference state. Subtracting the opposite experience keeps the preference's \emph{direction}. Subtracting a set of alternatives, or nothing, keeps its \emph{concept}. Because negation is weakly encoded \citep{jang2023negated}, a stored dislike is then recalled as a like: with the alternatives subtracted, the memory pushed every dislike item toward the thing the user dislikes (original judge), and the average effect pointed the wrong way on both models under every judge that passed a sanity check (Figure~\ref{fig:main}b).
\end{itemize}
The memory is nevertheless selective: about 98\% of unrelated replies are token-for-token identical to those with no memory, against 6--10\% when the gate is forced open.

Our contributions are: (1) a gated, training-free activation memory for user content, and its link to associative-memory model editing; (2) the fixed-average account and the three predictions it makes; (3) a pre-registered test of these predictions on two model scales, with controls and a cross-family judge panel; and (4) the reference effect, a concrete way in which a naive activation memory inverts what the user said. Together, the results support the procedural/declarative split assumed by \citet{zhao2026npm} and explain it: a pushed activation shift suits dispositions, while facts must be placed where attention can reach them: in context, as restored cached states, or in weights.

\section{Method}
\label{sec:method}

We read and modify the residual stream $h^{\ell}_t \in \mathbb{R}^d$ at the output of a few late decoder blocks $\ell$ of a frozen model. Everything below is done separately at each of these layers.

\paragraph{Experiences and moments.} An experience $e$ (e.g.\ \textit{``I can't stand jazz music.''}) is presented with a few ordinary follow-up turns $f$ (e.g.\ \textit{``I want to discover some new music.''}). Each follow-up is one \emph{moment}. We run each moment twice, once as $e$ followed by $f$ (\emph{with}) and once with a reference text in place of $e$, and we compare the states over the tokens of $f$. Tokens of the chat template that follow the user's text are excluded. In preliminary experiments, the choice of tokens to write did not matter (tokens selected by attention, by entropy or at random worked equally well), so a moment is stored as one pooled vector.

\paragraph{Shift and reference.} The shift of moment $j$ is a weighted mean over the follow-up's tokens,
\begin{equation}
  \Delta_j = \sum_{t \in f} w_t \bigl(h^{\text{with}}_t - h^{\text{ref}}_t\bigr), \qquad w_t \propto H_t,
\end{equation}
where $H_t$ is the model's next-token entropy at position $t$ in the run without the experience, and the weights sum to one. The \emph{reference} decides what $\Delta_j$ measures. We compare three:
\begin{itemize}
  \item \textbf{without}: the follow-up alone, so $\Delta$ is everything the experience added;
  \item \textbf{opposite}: the opposite experience (\textit{``I'm a huge jazz fan.''}), which cancels the shared topic and keeps the attitude;
  \item \textbf{centroid}: the mean state over eight same-frame alternatives (\textit{``I can't stand rock music.''}, \textit{``\ldots{} classical music.''}, \ldots), which cancels the frame and keeps what is specific to this experience. This is the construction used for concept vectors by \citet{lindsey2026introspection}.
\end{itemize}
Facts use \textit{without}. Preferences use \textit{opposite} unless stated otherwise.

\paragraph{Key.} Let $\mu$ be the mean state on 100 generic prompts and their continuations, and let $W = U_r(\Lambda_r + \epsilon \bar\lambda I)^{-1/2}$ whiten the top $r{=}256$ principal directions of those states ($\epsilon{=}0.01$, $\bar\lambda$ the mean eigenvalue). The key of moment $j$ is the normalised running mean of the whitened states over the follow-up's user tokens in the run without the experience,
\begin{equation}
  k_j = \operatorname{normalize}\Bigl(\tfrac{1}{|f|}\sum_{t\in f} W^{\top}(h_t - \mu)\Bigr) \in \mathbb{R}^{r}.
\end{equation}
The key thus describes the \emph{situation} the experience was filed under, not the experience itself. Whitening suppresses the components that all states share; without it, keys of unrelated situations were nearly as similar as keys of related ones.

\paragraph{Store.} A memory is a matrix $M \in \mathbb{R}^{d\times r}$ that maps keys to shifts. Its moments are written with the delta rule $M \leftarrow M + (\Delta_j - M k_j)\,k_j^{\top}$ of fast-weight memories \citep{schlag2021linear}, so each moment's own key recalls its shift exactly. When several memories share one store we use the least-squares solution (recursive least squares with ridge $\lambda{=}0.1$), which removes the dependence on write order. This is the closed-form key--value update of ROME and MEMIT \citep{meng2022rome,meng2023memit}, applied to activations rather than weights. In this paper each item has its own memory.

\paragraph{Read.} In a later session, the read key $q_t$ is the same pooled, whitened, normalised quantity over the user text seen so far. The gate opens when the best match with the stored keys exceeds a threshold $\theta$,
\begin{align}
  g_t &= \bigl[\max\nolimits_j \cos(q_t, k_j) > \theta\bigr], \\
  h_t &\leftarrow h_t + \alpha\, g_t\, M q_t ,
\end{align}
where $[\cdot]$ is 1 if true and 0 otherwise, $\theta$ is the 95th percentile of the match on the generic prompts, set per item and layer. The gate is never open on the system prompt or the template head. Keys and gates are computed in a first, memory-free pass, so an injection cannot change its own gate; the second pass adds the recalled shift at each layer. The gate is a conditional-steering rule in the sense of \citet{lee2025cast}, except that the condition is written from a single experience rather than learned from labelled prompts, and its threshold is calibrated on ordinary text rather than tuned. $\alpha$ is the only free strength parameter, applied equally at each layer.

\paragraph{Layers.} On Qwen3.5-2B (24 blocks) we read and write at block outputs 20, 21 and 23, chosen in a preliminary sweep over layers 4--23 on a subset of the items, in which three late layers together carried much more than any single layer (Appendix~\ref{app:prelim}). On Qwen3.5-9B (32 blocks) we use the same relative depths (27, 28, 31).

\section{Why a stored shift keeps some content and not other}
\label{sec:theory}

In context, an experience affects a later reply through attention. At each new position the model attends back to the tokens of the experience and computes what it needs for the question at hand. \citet{liu2024icv} make this explicit: the attention output is a mixture of the query's own state and the context's states, with weights that depend on the query. \citet{geva2023dissecting} show the same for facts: the attribute is not present at the subject's position in ready-made form; upper-layer attention extracts it, driven by the question. A direct test on in-context facts (Appendix~\ref{app:stage0}) finds the same: the answer reads the value from the position where it was stated, and the subject's position holds no copy of it.

A stored shift removes the recomputation. It is one vector per moment, averaged over the follow-up's tokens, and it is added unchanged at every position where the gate is open. Whatever part of the experience's effect varies with the question is averaged away; what is constant across questions remains. This gives three predictions.

\paragraph{P1: facts fail.} A fact matters at one slot (\textit{``your dog's name is \_\_\_''}). Its question-specific retrieval does not survive averaging, and what remains is the already-extracted output, \textit{say ``Pepper''}, which only exists near the end of the network. Pushed at every position, such a shift either does nothing or makes the model repeat the word; it should not be used correctly in a reply, and it cannot act as a premise, which requires combining the fact with the question.

\paragraph{P2: leanings survive, in a dose window.} A broad preference (\textit{``I'm vegetarian''}) pushes in the same direction for any food question, so a constant push helps. But the push is added at every position whether or not it is needed, so beyond some strength it should overwhelm the reply.

\paragraph{P3: the reference decides what is stored.} The shift is the state with the experience minus a reference state, so it keeps whatever differs between the two. Against the \emph{opposite} experience the topic cancels and the attitude remains. Against \emph{alternatives} in the same frame (\textit{``I can't stand rock / pop / \ldots''}) the frame cancels and the specific concept remains. Against \emph{nothing}, both remain. Because negation is weakly represented \citep{jang2023negated,lindsey2026introspection}, the concept dominates the last two, so a stored \emph{dislike} should be recalled as a \emph{like}. For positively stated preferences, concept and direction agree, so all three references should push the same way.

\section{Experimental setup}
\label{sec:setup}

\paragraph{Models.} Qwen3.5-2B (24 blocks) and Qwen3.5-9B (32 blocks), text path only, in fp32, with thinking disabled. The 2B ran on NVIDIA V100 (32\,GB) and the 9B on A100 (80\,GB) GPUs\ifdefined\REVIEW\else{} of the DelftBlue supercomputer \citep{DHPC2024}\fi. Like the rest of the Qwen3.5 family \citep{qwen2026omni}, both interleave linear-attention (Gated DeltaNet) and full-attention blocks.

\paragraph{Items.} 41 items in four groups, each with one experience sentence, three follow-up turns that write the memory, and later prompts (Appendix~\ref{app:items}):
\begin{itemize}
  \item 15 \textbf{leanings} (e.g.\ vegetarian, tight budget, early riser), each with its opposite and eight same-frame alternatives;
  \item 8 \textbf{dislikes} (e.g.\ \textit{``I can't stand jazz music.''}, \textit{``I don't drink alcohol.''});
  \item 6 \textbf{one-of-many} preferences (e.g.\ a home country, a favourite genre);
  \item 12 \textbf{facts} (names, job, home town, favourite colour, \ldots).
\end{itemize}
Preference items are probed with 1 related and 5 ambiguous open-ended prompts (\textit{``What should I listen to while I cook dinner?''}) and with balanced yes/no questions. Fact items are probed with a direct question, two \emph{use} prompts that need the fact without asking for it (\textit{``Write a short birthday message to my sister.''}), and balanced yes/no questions (\textit{``Is my dog called Pepper?''} / \textit{``\ldots{} Max?''}). 50 unrelated prompts test leakage.

\paragraph{Conditions.} No memory (\texttt{nomem}); the experience written into the prompt (\texttt{ctx}, the ceiling); the memory at $\alpha \in \{0.5, 1, 2, 3\}$ per layer; and four controls at $\alpha{=}2$: a random vector with the recall's norm through the item's own gate (\texttt{rand}), another item's shift under this item's gate (\texttt{swap}), a memory written from a neutral sentence (\texttt{placebo}), and the item's memory with the gate removed (\texttt{gate\_on}). For the mechanism (P3), the dislikes, one-of-many items and four leanings are also written with the \emph{centroid} and \emph{without} references at $\alpha \in \{0.5, 1, 2\}$.

\paragraph{Generation.} Each prompt gets one greedy and four sampled replies ($T{=}1$, top-$p$ 0.95, top-$k$ 20, presence penalty 1.5, at most 120 tokens). Every condition uses the same random streams, so differences come from the memory. A second seed set repeats the main conditions.

\paragraph{Measures.} Replies are labelled by an LLM judge \citep{zheng2023judging} (Qwen3.5-9B, thinking off) with fixed rubrics (Appendix~\ref{app:judge}). For preferences the primary measure is the \emph{judge lean}: $+1$ for a coherent reply whose recommendations fit the user's trait, $-1$ for a coherent reply against it, and $0$ otherwise. Incoherent replies count as 0, so loops and garbled text never count as success. For facts the primary measure is \emph{clean use} on the use prompts: the reply uses the correct value as a fact about the user and is coherent. We also report word-list measures, repetition loops (repeated 4-gram rate $\ge 0.3$), the judge's incoherence rate, yes/no accuracy and margins, and the logit lens \citep{nostalgebraist2020logitlens} of stored shifts. Every number has a 95\% bootstrap confidence interval over items and prompts, paired across conditions.

\paragraph{Pre-registration and judge robustness.} Claims, decision rules, an item-inclusion rule (an item is kept if its in-context effect is at least 0.3) and the statistics were committed before any run (Appendix~\ref{app:prereg}). Because the original judge is from the same family as the models it grades, and LLM judges can favour their own outputs \citep{zheng2023judging}, we re-labelled all 34{,}200 replies with three judges from other families (gpt-oss-20B, Gemma-4-12B, Llama-3.1-8B) and a larger Qwen model, and recomputed every judge-based verdict under each judge and under majority votes (Appendix~\ref{app:judge}). This panel was added after the pre-registered analysis.

\section{Results}
\label{sec:results}

\subsection{The memory is selective}
With the gate, 98.4\% (2B) and 98.2\% (9B) of replies to the 50 unrelated prompts are token-for-token identical to those with no memory, at $\alpha{=}2$. With the gate removed, only 5.7\% and 10.1\% are; the KL divergence from the no-memory distribution rises four- to six-fold (Figure~\ref{fig:sel} in the appendix). The gate opens somewhere on 3.5\% and 4.0\% of unrelated prompts. Selectivity therefore does not explain any of the failures below: when the memory misbehaves, it does so where it is supposed to fire.

\subsection{Facts are stored as a late word and are not used}

\begin{figure}[t]
  \centering
  \includegraphics[width=\columnwidth]{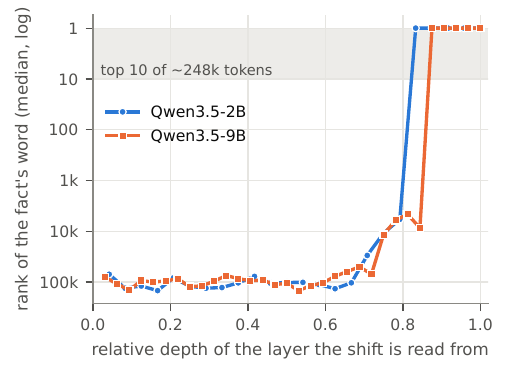}
  \caption{The stored fact shift read through the output layer at every block (logit lens). The median rank of the fact's word (e.g.\ \textit{Pepper}) among about 248k tokens stays near 100k until the last sixth of the network, then jumps to the top.}
  \label{fig:lens}
\end{figure}

\begin{figure*}[t]
  \centering
  \includegraphics[width=0.92\textwidth]{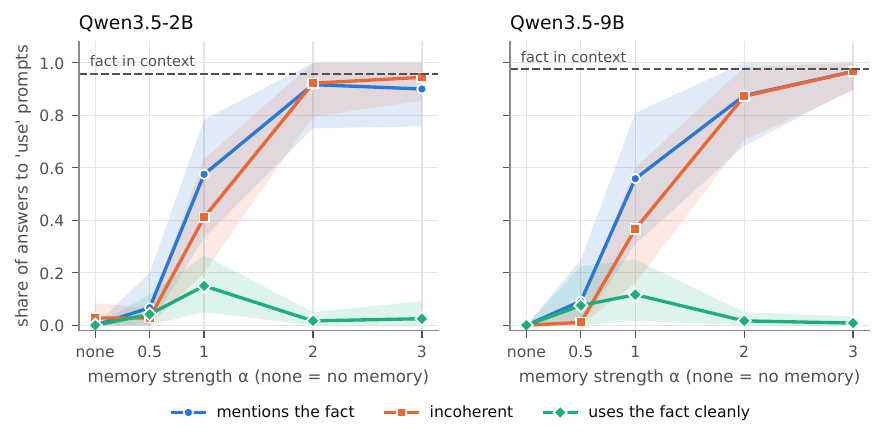}
  \caption{Fact memories on the use prompts. As the strength grows, more replies mention the fact (blue) and more are incoherent (orange), but replies that use the fact cleanly (green, judge) stay at or below 15\%. With the fact in context, 96--98\% use it cleanly (dashed). Shaded: 95\% CIs over items and prompts.}
  \label{fig:facts}
\end{figure*}

\paragraph{What is stored.} Read through the output layer, the stored fact shift ranks the fact's word in the top 10 of the vocabulary only at relative depth $\ge 0.83$ (2B) and $\ge 0.875$ (9B); at half depth its median rank is about 70{,}000 on both models (Figure~\ref{fig:lens}). The shift encodes the word to say, and only where saying it is decided. This was pre-registered as C6 and holds on both models. The logit lens is only approximate before the last layers \citep{belrose2023tuned}, but the behavioural results below point the same way.

\paragraph{What happens.} Clean use of the fact stays far below the in-context ceiling at every strength (Figure~\ref{fig:facts}): at most 0.15 (2B) and 0.12 (9B), at $\alpha{=}1$, against 0.96 and 0.98 in context. Increasing the strength does not convert mentions into use. At $\alpha{=}2$, 89\% (2B) and 86\% (9B) of replies mention the fact without using it cleanly, mostly as loops (82\% on both). A typical reply to \textit{``What's my dog's name?''} at $\alpha{=}2$ is \textit{``I Pepper Pepper Pepper\ldots''}; in context it is \textit{``Your dog's name is Pepper.''} At lower strength the model usually answers that it does not know. The three controls never produce the target. This was pre-registered as C3 and holds on both models and under every judge that passed our sanity check (Appendix~\ref{app:judge}).

\paragraph{No premises.} On balanced yes/no questions about the facts, the models without memory answer ``No'' to everything (balanced accuracy 0.50); in context they reach 0.94. The memory never improves the balanced accuracy at any strength, on either model. On the 2B, at $\alpha{=}2$, it even lowers it (by 0.15, CI $[-0.27, -0.04]$), which formally fails the pre-registered rule for C4, whose wording required no change. The yes/no margins move by a general tilt rather than toward the correct answer: on the 9B at $\alpha{=}2$ the Yes-correct margin rises by 0.78 and the No-correct margin falls by 0.73. The same holds for the preference items. We read this as no evidence for premises on either model, with the 2B result being a decrease rather than a gain.

\subsection{Leanings transfer only in a narrow window}

\begin{figure*}[t]
  \centering
  \includegraphics[width=0.92\textwidth]{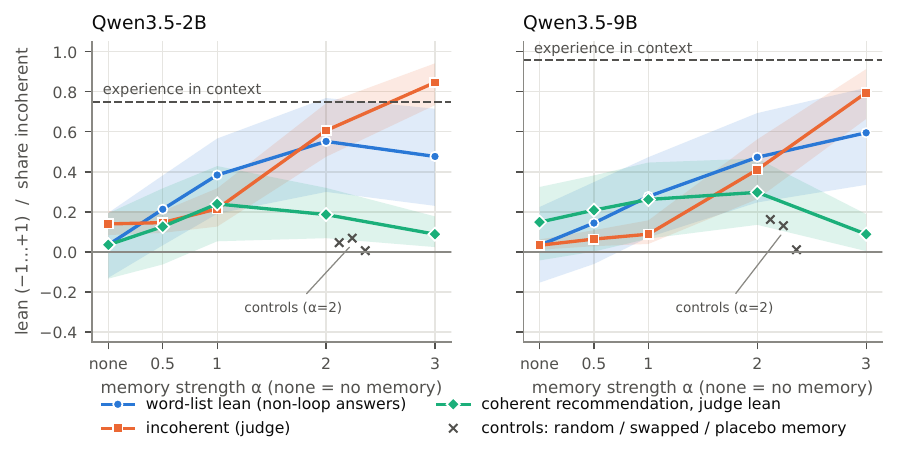}
  \caption{Leaning memories (15 items) under the original judge. The word-list lean of non-looping replies (blue) keeps rising with strength, but the share of incoherent replies (orange) rises faster, so the judge lean of coherent recommendations (green) peaks at $\alpha{\approx}1$--2 and falls. Crosses: controls at $\alpha{=}2$. Dashed: the experience in context.}
  \label{fig:leanings}
\end{figure*}

At $\alpha{=}1$, leaning memories shift open-ended replies in the right direction: the judge lean rises by 0.20 (CI $[0.04, 0.38]$, 2B) and 0.11 ($[0.004, 0.24]$, 9B) over no memory, against 0.72 and 0.81 for the experience in context (Figure~\ref{fig:leanings}). An independent second set of samples reproduces these values within about 0.03. At $\alpha{=}2$ the word-list lean keeps rising, but the judge calls 61\% (2B) and 41\% (9B) of replies incoherent, against 14\% and 3\% without memory, and the gain's interval includes zero. At $\alpha{=}3$ most replies are incoherent. The random-vector control's gain is near zero, but its paired difference from the memory at $\alpha{=}2$ also includes zero. The pre-registered claim C2, which required a gain at both $\alpha{=}1$ and $\alpha{=}2$ and a clear separation from every control, therefore \emph{fails} on both models.

This failure depends on the judge. The original judge is the strictest about coherence: it labels 62\% of preference replies coherent, against 72--76\% for the other competent judges. Under a majority vote of the three cross-family judges, the $\alpha{=}2$ gain is 0.36 ($[0.15, 0.58]$, 2B) and 0.28 ($[0.10, 0.48]$, 9B), and the controls separate. The effect is also uneven across items: strong for \textit{early riser}, \textit{prefers quiet}, \textit{vegan} and \textit{vegetarian}, absent for \textit{celiac}, \textit{lactose intolerant} and \textit{has young kids}. We conclude that leanings do transfer, but only within a narrow strength window, with a size that depends on how strictly incoherence is judged. Single-item preferences (one-of-many) barely move at any strength (gain at most 0.14).

\subsection{What the shift subtracts decides concept or direction}

\begin{figure*}[t]
  \centering
  \includegraphics[width=0.92\textwidth]{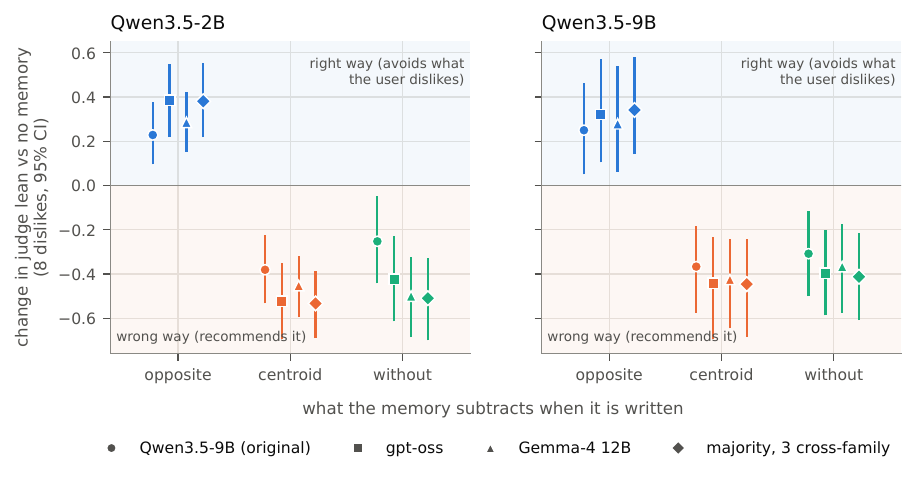}
  \caption{Dislike memories (2B: 7 items kept by the inclusion rule; 9B: 8), written with each reference and compared at the highest strength with at most 10\% loops ($\alpha{=}1$ for all). Change in judge lean against no memory: positive means the replies avoid what the user dislikes. Only the opposite reference moves replies the right way; the centroid and \emph{without} references push toward the disliked thing, under every competent judge. Bars: 95\% CIs.}
  \label{fig:dislikes}
\end{figure*}

For dislikes, the three references move replies in opposite directions (Figure~\ref{fig:dislikes}). Compared at the same loop rate, the memory written against the \textbf{opposite} moves replies away from the disliked thing, by $+0.23$ ($[0.05, 0.44]$, 2B) and $+0.25$ ($[0.02, 0.50]$, 9B). The memory written against the \textbf{centroid} moves them toward it, by $-0.38$ ($[-0.57, -0.19]$) and $-0.37$ ($[-0.59, -0.15]$), and does so for every item on both models. The memory written against \textbf{nothing} also moves them toward it, by $-0.25$ and $-0.31$, for 5 of 7 and 7 of 8 items. Asked for music to cook to after \textit{``I can't stand jazz music.''}, the 9B with the centroid memory recommends \textit{``Jazz Classics: Miles Davis, Bill Evans, or Chet Baker''} (Figure~\ref{fig:main}b). This was pre-registered as C5 and holds on both models. It also holds under gpt-oss, Gemma and both majority votes, with larger effects ($-0.42$ to $-0.53$ for the centroid). The same-family Qwen3.6 judge agrees on every sign, but its \emph{without} interval touches zero.

The logit lens shows the cause directly (Figure~\ref{fig:lensdis} in the appendix). For dislikes, the stored shift's preference for liked-side over disliked-side words at the last layer is $+3.7$ (2B) and $+4.2$ (9B) for the opposite reference, but $-10.6$ and $-11.6$ for the centroid and $-3.2$ and $-3.6$ for \emph{without}: those shifts point at the disliked thing. For positively stated leanings, all three references point the same way ($+4.8$ to $+9.1$), and their behavioural effects agree in sign, as P3 predicts.

\section{Related work}
\label{sec:related}

\paragraph{Steering and its limits.} Adding vectors to the residual stream controls dispositions such as honesty, refusal and persona \citep{turner2023actadd,zou2023repe,rimsky2024caa,cao2024bipo}. Its effects vary across prompts and concepts \citep{tan2024reliability,braun2025unreliability}, suit traits that shape a whole reply better than local ones \citep{subbiah2026limits}, and trade expression for coherence \citep{subbiah2026limits,suri2025memorization}. Injected concepts overwhelm the model at high strength, are read as the model's own, and are not removed by negation \citep{lindsey2026introspection}. Our leaning and fact results are consistent with all of these; we add the memory setting and the role of the reference.

\paragraph{Conditional and personalised steering.} CAST gates a steering vector on the similarity between the current state and a learned condition \citep{lee2025cast}; our gate is the same idea with the condition written from a single experience. In-context vectors compress demonstrations into a vector \citep{liu2024icv}, and context steering amplifies the effect of a user's context on the output probabilities \citep{he2025cos}; both need the context present. BiPO trains a personalised steering vector for a chosen behaviour \citep{cao2024bipo}. Context steering notes, informally, that a single context also makes its topic more likely and that a contrasting pair isolates the attribute \citep{he2025cos}; we show the consequence for stored memories systematically.

\paragraph{Memory in activations and weights.} Neural Procedural Memory stores an agent's skills as retrieved steering vectors and leaves declarative content to text \citep{zhao2026npm}; our results support that division and explain it. Facts can be written usefully into weights by key--value updates at the subject's position in middle-layer MLPs \citep{meng2022rome,meng2023memit}; our store uses the same update in activations, without optimising the stored value. Factual recall itself relies on question-driven attention in upper layers \citep{geva2023dissecting}, which a stored shift cannot reproduce. Retrieval-augmented generation and agent memory systems keep facts in context \citep{lewis2020rag,packer2023memgpt}. A related line keeps facts as cached \emph{states}: retrieving stored key--value blocks into attention \citep{xiao2024infllm,fountas2025emllm}, reusing cached chunks across prompts and positions \citep{gim2024promptcache,yao2025cacheblend,hu2025epic}, or training a model to attend to explicit or sparse key--value memories \citep{yang2024memory3,wang2025kblam}. All of these keep content at positions attention can reach, which is the property a pushed shift lacks. In context, attributes are bound to entities and fetched by position \citep{feng2024binding,prakash2024entity,prakash2026lookbacks}, so content must sit at a position to be bound. Our store is also a fast-weight memory, the same delta-rule mechanism that linear-attention layers implement \citep{schlag2021linear}. A trained version of that design, a delta-rule fast-weight memory read through a gated residual addition, does carry facts \citep{zhao2026fwpkm}, because the read path is trained with the model; a frozen model was never trained to read a pushed shift.

\section{Conclusion}
\label{sec:conclusion}

A gated activation memory beside a frozen model can carry broad leanings into a later session, selectively and only within a narrow strength window. It cannot carry facts or premises as a pushed shift: a stored fact is the word to say, present only at the end of the network. And the way the memory is written determines what it means: unless the opposite experience is subtracted, a stored dislike is recalled as a like. We read all three results as consequences of one fact: a stored shift is a fixed average of an effect that attention recomputes, in context, for each question. A pushed activation shift is therefore a natural home for dispositions, while facts must be placed where attention can reach them: in context, as restored cached states, or in weights. A practical next step is to store the concept and its sign separately, so that the direction of a preference no longer depends on a hand-written opposite; follow-up experiments find that this works when the sign is known, while reading the sign automatically remains open.

\section*{Limitations}
\label{sec:limits}

\paragraph{Scale and items.} Two models from one family (2B and 9B) and 41 items, with 6--15 per group. The dislike result rests on 7--8 items. Larger models or other families may behave differently.

\paragraph{Judges.} The primary labels come from an LLM judge from the same family as the graded models. The cross-family panel shows that C3 and C5 do not depend on the judge, but not that the labels are correct; no human labels were collected. Llama-3.1-8B was excluded from the panel after the fact because it failed an objective check (Appendix~\ref{app:judge}); its verdicts are reported. The panel itself was not pre-registered.

\paragraph{Design.} Each memory is written from three hand-written follow-ups, and the opposite and centroid references are hand-written. A practical system would need these references without human input; in follow-up experiments (reported separately), references generated by a larger model of the same family reproduce the reference effect. We test isolated memories only; with many memories in one store, interference may change the picture. The theoretical account in Section~\ref{sec:theory} is an interpretation that fits the results, not a proof.

\section*{Ethical Considerations}
The memory studied here stores information about users inside a model's activations. Such stores raise the same privacy questions as any user memory (consent, inspection and deletion), with the added difficulty that an activation shift is not human-readable. Our results also show two failure modes that matter for deployment: a naive memory can recall a dislike as a like, and an injected memory can be presented by the model as its own trait. All experiments use synthetic user statements written by the author; no personal data was collected.

\ifdefined\REVIEW\else
\section*{Acknowledgements}
Experiment code, analysis scripts and the drafting of this paper were assisted by Claude (Anthropic). The author designed the study and the pre-registration, and checked all reported results. The author acknowledges the use of computational resources of the DelftBlue supercomputer, provided by Delft High Performance Computing Centre (\url{https://www.tudelft.nl/dhpc}).
\fi

\bibliography{refs}

\appendix

\section{Items and prompts}
\label{app:items}

\begin{table}[h]
\centering\small
\begin{tabular}{@{}lp{0.68\columnwidth}@{}}
\toprule
Group & Items \\
\midrule
Leanings (15) & vegetarian, vegan, lactose intolerant, celiac, diabetes, loves hiking, tight budget, early riser, prefers quiet, eco-conscious, has young kids, loves cooking, loves cold, bookworm, history buff \\
Dislikes (8) & jazz, alcohol, horror films, spicy food, hiking, seafood, the beach, sweet food \\
One-of-many (6) & loves jazz, lives in Norway, lives in Singapore, Japanese food, tennis, horror movies \\
Facts (12) & dog's, cat's, sister's, brother's, grandmother's, best friend's and boss's names; job; favourite colour; home town; first language; university \\
\bottomrule
\end{tabular}
\caption{The 41 items. Each has an experience sentence (e.g.\ \textit{``I can't stand jazz music.''}), an opposite (\textit{``I'm a huge jazz fan.''}), three follow-up turns and its probes; preferences also have eight same-frame alternatives for the centroid. The pre-registered inclusion rule removed one item (\textit{hates horror}) on the 2B.}
\end{table}

\section{Judges}
\label{app:judge}

\paragraph{Rubrics.} The preference rubric asks whether the reply's recommendations fit the user's trait (\emph{toward}), go against it (\emph{away}) or neither (\emph{neutral}), explicitly judging what is recommended rather than which words are mentioned; whether the reply is coherent; and whether the assistant claims the trait as its own. The fact rubric asks whether the reply uses the correct value as a fact about the user (a mention inside garbled or repetitive text does not count), whether it is coherent, whether the assistant claims the fact as its own, and whether it asserts a wrong value. Full prompts are in \codeavail.

\paragraph{Panel.} All 34{,}200 replies were re-labelled by gpt-oss-20B, Gemma-4-12B, Llama-3.1-8B and Qwen3.6-35B-A3B (same family; reported, not voting), with the same prompts, temperature 0 and JSON-constrained output; fewer than 0.1\% of outputs failed to parse. Table~\ref{tab:panel} gives each claim's verdict under each judge.

\begin{table*}[t]
\centering\small
\begin{tabular}{@{}llccccccc@{}}
\toprule
Claim & Model & Original & gpt-oss & Gemma & Llama$^\dagger$ & Qwen3.6$^\ddagger$ & Maj.\ (orig.+3) & Maj.\ (3 cross) \\
\midrule
C2 leanings & 2B & no & no & no & no & yes & yes & yes \\
            & 9B & no & no & no & no & no & yes & yes \\
C3 facts    & 2B & yes & yes & yes & no & yes & yes & yes \\
            & 9B & yes & yes & yes & no & yes & yes & yes \\
C5 dislikes & 2B & yes & yes & yes & no & no & yes & yes \\
            & 9B & yes & yes & yes & no & no & yes & yes \\
\bottomrule
\end{tabular}
\caption{Does each pre-registered claim hold under each judge? $^\dagger$Fails the sanity check below and is excluded from our conclusions. $^\ddagger$Same family as the original judge; its C5 signs all agree, but its \emph{without} interval touches zero.}
\label{tab:panel}
\end{table*}

\paragraph{Sanity check.} A reply cannot use a name it does not contain. Among replies a judge labelled as using the fact, the target string was absent in 1\% for Qwen3.6, Gemma and gpt-oss, but in 48\% for Llama-3.1-8B, which labels 92\% of all fact replies as using the fact (others: 16--26\%) and agrees with the other judges on this label at chance level ($\kappa \approx 0.05$). We therefore exclude it. Among the remaining judges, agreement on coherence is high (Fleiss' $\kappa = 0.90$ on facts) and the original judge agrees with Gemma on fact use at $\kappa = 0.93$.

\section{Preliminary experiments}
\label{app:prelim}

The design choices were fixed in earlier, smaller experiments that were not pre-registered. On Qwen2.5-1.5B, unwhitened per-token keys made the memory fire on unrelated prompts; whitened pooled keys with a calibrated threshold made unrelated replies identical to no memory. Selecting which tokens to write by attention, by entropy or at random made no difference. On Qwen3.5-2B, a sweep over layers 4--23 found fact-specific effects only in the last five layers, and three late layers together carried much more than any single one. An earlier version of the reference comparison (11 preference items) first showed the dislike reversal that core experiment C5 tests.

\section{Additional figures}
\label{app:figs}

\begin{figure}[h]
  \centering
  \includegraphics[width=\columnwidth]{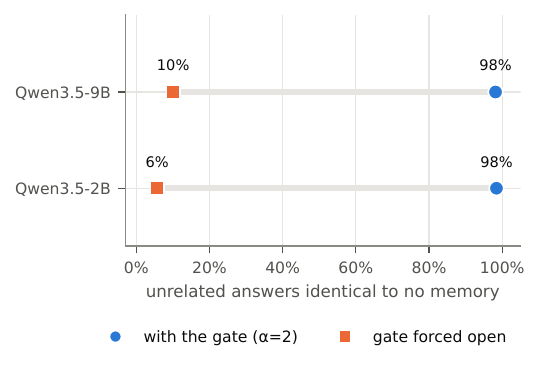}
  \caption{Unrelated replies identical to no memory, at $\alpha{=}2$, with the gate and with the gate forced open.}
  \label{fig:sel}
\end{figure}

\begin{figure*}[h]
  \centering
  \includegraphics[width=0.92\textwidth]{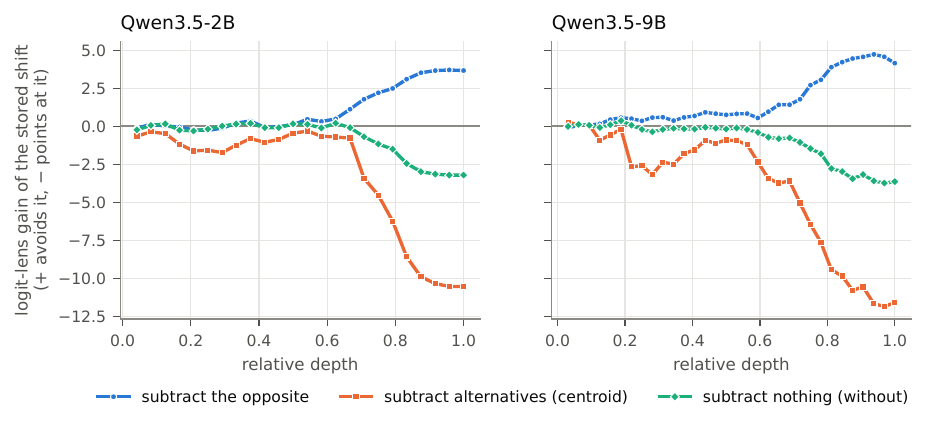}
  \caption{Logit lens of the stored dislike shifts at every block: mean logit of words on the liked side minus the disliked side (positive: avoids the disliked thing). The centroid and \emph{without} shifts point at the disliked thing in the late layers.}
  \label{fig:lensdis}
\end{figure*}

\FloatBarrier
\section{Where an in-context fact is read from}
\label{app:stage0}

Section~\ref{sec:theory} explains why facts fail with the account of \citet{geva2023dissecting}, which concerns facts stored in the weights. We also measured the in-context case directly, in a follow-up experiment that was pre-registered separately and run after the main study; it is reported here only as supporting evidence for P1.

\paragraph{Setup.} The fact is in context (\textit{``My dog's name is Pepper. What's my dog's name?''}), and the answer is read at the teacher-forced prefix \textit{``Your dog's name is''}. We run the same prompt with a counter value (\textit{Rocky}) in an identical token layout and patch the residual stream of one block, or a band of five blocks, from the true-value run into the counter-value run at one site: the subject token of the question (\textit{``dog''}, after the name) or the attribute's own position (\textit{``Pepper''}). The read-out is the share of the flip in $\mathrm{logit}(\textit{Pepper}) - \mathrm{logit}(\textit{Rocky})$ that the patch recovers, averaged over the 7 primary fact items with 95\% bootstrap CIs over items. On Qwen2.5-1.5B we also remove attention from the answer position to each site over windows of five layers (attention knockout); Qwen3.5-2B's Gated DeltaNet blocks fold every past token into one state, so no single edge can be removed there.

\begin{table}[h]
\centering\scriptsize
\setlength{\tabcolsep}{2.5pt}
\begin{tabular}{@{}lcc@{}}
\toprule
 & Qwen2.5-1.5B & Qwen3.5-2B \\
\midrule
\multicolumn{3}{@{}l}{\emph{Patching (share of flip recovered, best cell)}} \\
subject (\textit{dog}) & 0.00 [0.00, 0.00] & 0.00 [0.00, 0.01] \\
attribute (\textit{Pepper}) & 1.00 [1.00, 1.00] & 1.00 [1.00, 1.01] \\
\midrule
\multicolumn{3}{@{}l}{\emph{Knockout of the answer's attention (share of gain lost)}} \\
to subject, best window & 0.00 [0.00, 0.00] & -- \\
to attribute, last window & 0.59 [0.41, 0.81] & -- \\
\bottomrule
\end{tabular}
\caption{Share of the true-vs-counter logit flip recovered by patching one site, and share of the in-context gain lost by blocking attention to it. Patching the subject recovers nothing at any block or band on either model; patching the attribute recovers everything up to the late blocks. Blocking the answer's attention to the attribute costs 0.59 only in the last window (layers 23--27); the rest flows through relay tokens.}
\label{tab:stage0}
\end{table}

\paragraph{Result.} Patching the question's subject token recovers 0.00 of the flip at every block and band on both models (Table~\ref{tab:stage0}). Patching the attribute's own position recovers 1.00 up to the late blocks: on Qwen2.5-1.5B it falls only in the last blocks, and on Qwen3.5-2B it drops sharply at block 19 of 24. Blocking the answer's attention to the subject costs nothing in any window; blocking its attention to the attribute costs 0.59 [0.41, 0.81] of the in-context gain, and only in the last window. An exploratory check of the subject mention in the answer prefix (\textit{``Your \textbf{dog}'s name is''}) also recovers at most 0.03. The answer therefore reads the value from the position where it was stated, late, and the subject's position holds no copy of it (it may still carry a pointer in the sense of \citet{prakash2026lookbacks}). A vector pushed at other positions has nothing to imitate.

\paragraph{Caveats.} These are different models from the main runs (Qwen2.5-1.5B and Qwen3.5-2B, not the 9B), 7 items, and a later pre-registration.

\section{Pre-registration and reproduction}
\label{app:prereg}

The claims (C1 selectivity, C2 leanings, C3 facts, C4 no premises, C5 reference, C6 a late word), the primary measures, the decision rules, the item-inclusion rule and the bootstrap were committed before any run (\texttt{experiments/core\_v1/PREREG.md}). Results against the pre-registered rules: C1, C3, C5 and C6 hold on both models; C2 fails on both; C4 holds on the 9B and fails on the 2B (a decrease in accuracy, not a gain). Deviations: the judge panel and the exclusion of Llama-3.1-8B were added after the pre-registered analysis; the pre-registered human audit of the judge was not carried out. All code, configurations, seeds and the commands that regenerate every table and figure from the saved outputs are available in \codeavail.

\end{document}